\documentclass[sigconf]{acmart}

\makeatletter
\def\@ACM@checkaffil{
    \if@ACM@instpresent\else
    \ClassWarningNoLine{\@classname}{No institution present for an affiliation}%
    \fi
    \if@ACM@citypresent\else
    \ClassWarningNoLine{\@classname}{No city present for an affiliation}%
    \fi
    \if@ACM@countrypresent\else
        \ClassWarningNoLine{\@classname}{No country present for an affiliation}%
    \fi
}
\makeatother

\usepackage{amsmath}
\usepackage{enumitem}
\usepackage{subcaption}
\usepackage{multirow}
\usepackage{array}
\usepackage{dsfont}
\usepackage{mathtools}
\newcolumntype{P}[1]{>{\centering\arraybackslash}p{#1}}
\usepackage{amsmath, bm}
\usepackage{enumitem}
\usepackage{booktabs}
\usepackage{array, makecell}
\usepackage{booktabs}
\usepackage{placeins}
\usepackage{hhline}

\newlength{\ColWidthNormal} 
\newlength{\ColWidthRowHeader} 
\newlength{\RuleOffsetLeft} 
\newlength{\RuleThicknessNormal} 
\newcolumntype{C}{>{\centering\arraybackslash\leavevmode}p{\ColWidthNormal}}

\AtBeginDocument{%
  \providecommand\BibTeX{{%
    \normalfont B\kern-0.5em{\scshape i\kern-0.25em b}\kern-0.8em\TeX}}}

\setcopyright{acmcopyright}
\copyrightyear{2018}
\acmYear{2018}
\acmDOI{XXXXXXX.XXXXXXX}

\acmConference[Conference acronym 'XX]{Make sure to enter the correct
  conference title from your rights confirmation email}{June 03--05,
  2018}{Woodstock, NY}
\acmBooktitle{Woodstock '18: ACM Symposium on Neural Gaze Detection,
 June 03--05, 2018, Woodstock, NY} 
\acmPrice{15.00}
\acmISBN{978-1-4503-XXXX-X/18/06}

\copyrightyear{2023}
\acmYear{2023}
\setcopyright{acmlicensed}\acmConference[SIGIR '23]{Proceedings of the 46th International ACM SIGIR Conference on Research and Development in Information Retrieval}{July 23--27, 2023}{Taipei, Taiwan}
\acmBooktitle{Proceedings of the 46th International ACM SIGIR Conference on Research and Development in Information Retrieval (SIGIR '23), July 23--27, 2023, Taipei, Taiwan}
\acmPrice{15.00}
\acmDOI{10.1145/3539618.3591686  }
\acmISBN{978-1-4503-9408-6/  23/07}

\begin{document}

\title{BiTimeBERT: Extending Pre-Trained Language Representations with Bi-Temporal Information}

\author{Jiexin Wang}
\affiliation{%
  \institution{South China University of Technology, China}
  }
\email{wangjiexin.scut@yahoo.com}

\author{Adam Jatowt}
\affiliation{%
  \institution{University of Innsbruck, Austria}
  }
\email{adam.jatowt@uibk.ac.at}

\author{Masatoshi Yoshikawa}
\affiliation{%
  \institution{Kyoto University, Japan}
  }
\email{yoshikawa@i.kyoto-u.ac.jp}

\author{Yi Cai}
\affiliation{%
  \institution{South China University of Technology, China}
  }
\email{ycai@scut.edu.cn}

\renewcommand{\shortauthors}{Wang, et al.}

\begin{abstract}

Time is an important aspect of documents and is used in a range of NLP and IR tasks. In this work, we investigate methods for incorporating temporal information during pre-training to further improve the performance on time-related tasks. Compared with common pre-trained language models like BERT which utilize synchronic document collections (e.g., BookCorpus and Wikipedia) as the training corpora, we use long-span temporal news article collection for building word representations. We introduce BiTimeBERT, a novel language representation model trained on a temporal collection of news articles via two new pre-training tasks, which harnesses two distinct temporal signals to construct time-aware language representations. The experimental results show that BiTimeBERT consistently outperforms BERT and other existing pre-trained models with substantial gains on different downstream NLP tasks and applications for which time is of importance (e.g., the accuracy improvement over BERT is 155\% on the event time estimation task).\footnote{The code is available at \url{https://github.com/WangJiexin/BiTimeBERT}.}
 
\end{abstract}



\maketitle

\section{Introduction}
Temporal signals constitute significant features in various types of text documents such as news articles or biographies. They can be leveraged to understand
chronology, causalities, developments, and ramifications of events, being helpful in a range of different NLP tasks. 
Utilizing temporal signals in information retrieval has received considerable attention recently, too. For example, researchers have addressed time-sensitive queries in search leading to the formation of a subset of Information Retrieval called Temporal Information Retrieval \cite{campos2015survey, kanhabua2016temporal} in which both query and document temporal aspects are of key concern. Event detection and ordering \cite{strotgen2012event,bookdercz}, timeline summarization \cite{alonso2009clustering,tran2015timeline,martschat2018temporally,steen2019abstractive, campos2021automatic}, event occurrence time prediction \cite{wang2021event}, temporal clustering \cite{campos2012disambiguating}, question answering \cite{pasca,wang2020answering} 
 and semantic change detection \cite{rosin2022temporal, rosin2022time} are other example tasks where utilizing temporal information has proven beneficial. 

Pre-trained language models such as BERT \cite{devlin2018bert}, RoBERTa \cite{liu2019roberta}, GPT \cite{radford2019language, brown2020language} have recently achieved impressive performance on a variety of downstream tasks, and have been commonly utilized for representing, evaluating or generating text. However, despite their great success, they still suffer from difficulty in capturing important information in domain-specific scenarios, as, typically, these models tend to be trained on large-scale general corpora (e.g., English Wikipedia) while their training is not adapted to the characteristics of documents in particular domains. For example, they are incapable of utilizing temporal signals like document timestamp, despite temporal information being of key importance for many tasks such as ones that involve processing news articles.

In this paper, we propose a novel, pre-trained language model called BiTimeBERT, which is trained on a temporal news collection by exploiting two key temporal aspects: \emph{document timestamp} and \emph{content time}, the latter being represented by temporal expressions embedded in news articles. 
In the recent years, exploiting these two kinds of temporal information in documents and queries has been gaining increased importance in IR and NLP. Their interplay can be utilized to develop time-specific search and exploration applications \cite{alonso2011temporal, campos2015survey, kanhabua2016temporal}, such as temporal web search \cite{stack2006full}, temporal question answering \cite{wang2020answering, wang2021improving}, search results diversification \cite{berberich2013temporal, styskin2011recency} and clustering \cite{alonso2009clustering, svore2012creating}, 
summarization \cite{barros2019natsum}, event ordering \cite{honovich2020machine}, etc.

While BiTimeBERT has been continually pre-trained with only very few computation resources (less than 80 GPU hours), it outperforms other language models by a large margin on several tasks. Moreover, with only a small size of task-specific training data (e.g., 20\% for the EventTime dataset in year granularity), it can achieve performance similar to the one of baselines that use entire data. 
To sum up, we make the following contributions in this work:  
\begin{enumerate}[leftmargin=1.5em, labelwidth=!]
\item We investigate the effectiveness of incorporating temporal information into pre-trained language models using three different pre-training tasks, and we demonstrate that injecting such information via specially designed time-oriented pre-training can improve performance in various downstream time-related tasks.
\item We propose a novel pre-trained language representation model called BiTimeBERT, which is trained through two new pre-training tasks that involve two kinds of temporal information (timestamp and content time). To our best knowledge, this is the first work to investigate both types of temporal signals when constructing language models. 
\item We conduct extensive experiments on diverse time-related tasks on 7 datasets that involve the two temporal aspects of text. The results demonstrate that BiTimeBERT achieves a new SOTA performance and can offer effective time-aware representations, thus it has the capability to be successfully used in applications for which time is crucial.
\end{enumerate}

\vspace{-0.9em}
\section{Related Work}
\vspace{-0.2em}
\subsection{Language Models for Specific Domains}
The problem with the generic language models like BERT and GPT is that they are pre-trained on general-purpose large-scale text corpora (e.g., Wikipedia), which is not effective for
applications on specific domains or particular tasks.
Some studies thus adapt pre-trained models to specific domains by directly applying the two pre-training tasks of BERT on domain-specific datasets. The well-known examples are SciBERT \cite{beltagy2019scibert} trained on scientific corpus, BioBERT \cite{lee2020biobert} generated using a biomedical document corpus, and ClinicalBERT \cite{huang2019clinicalbert} derived from a clinical corpus. Another line of work attempts to continually pre-train the available language models to target applications or tasks. For example, \citet{ke2019sentilare} propose SentiLARE for sentiment analysis task, which continually pre-trains RoBERTa model with the proposed label-aware masked language model on a sentiment analysis dataset. 
In another strain of work, \citet{xiong2019pretrained} design WKLM (Weakly Supervised Knowledge-Pretrained Language Model) for entity-related tasks conducting continual pre-training on a BERT model with the entity replacement objective. This objective requires the model to make a binary prediction indicating whether an entity has been replaced or not. The experimental results with WKLM suggest that this kind of adaptation can better capture knowledge about real-world entities. 

\vspace{-0.5em}
\subsection{Incorporating Time with Language Models}
In recent years, incorporating time with language models has also been investigated \cite{giulianelli2020analysing, dhingra2021time, cole2023salient, rosin2022temporal, rosin2022time}. 
\citet{dhingra2021time} propose a simple modification to pre-training that parametrizes masked language modeling (MLM) objective with timestamp information using temporally-scoped knowledge, and test the proposed language model on question answering. \citet{cole2023salient} introduce Temporal Span Masking task (TSM), a variant of Salient Span Masking (SSM) \cite{guu2020retrieval}. TSM, which involves masking the temporal expressions in sentences and training the model to generate them, is designed for enhancing the model's temporal understanding capabilities. These two models adopt Transformer encoder-decoder architectures, while most existing works are mainly based on Transformer encoder-only models for facilitating the combination of the temporal information. Additionally, the proposed encoder-based models mainly solve the task of semantic change detection that requires identifying which words underwent semantic drift and to what extent. 
\citet{giulianelli2020analysing} propose the first unsupervised approach to tackle the task by using contextualized embeddings from BERT. \citet{rosin2022temporal} extend the canonical self-attention \cite{vaswani2017attention} by incorporating timestamp information, which is used to compute attention scores. 
\citet{rosin2022time} introduce TempoBERT, a time-aware BERT model by preprocessing input texts to concatenate with the timestamp information, and then masking these tokens while training. Their solution achieves SOTA performance on semantic change detection. 
Although \citet{rosin2022time} additionally experiment with the sentence time prediction task,\footnote{In Section 5.5, we compare our proposed model with TempoBERT on both semantic change detection and sentence time prediction tasks.} they test TempoBERT on two datasets that are of rather coarse granularity, i.e., the number of classes under year granularity is 40, while it is 4 in the easier setting of a decade granularity. Moreover, the authors observed a small degradation in performance on both datasets of the sentence time prediction task when compared with the fine-tuned BERT model. 

Thus, as we see, the existing time-aware language models (except \cite{dhingra2021time, cole2023salient}) mainly focus on the problem of lexical semantic change detection. Nonetheless, typically pre-training corpora designed for semantic change detection are based on a sentence level, such that each data instance is a short sentence that also rarely contains any content time expression. Hence, existing language models for semantic change detection neglect either the content temporal information or the timestamp information, and might also lack generalization abilities to other time-related tasks which require long contents as input. In addition, because the timestamps of the pre-training corpora are at year granularity, the timestamp information can only be utilized at coarse granularity (i.e., year or even a decade).

Similar to the above pre-trained models (e.g., TempoBERT \cite{rosin2022time}), our proposal is also a Transformer-based \cite{vaswani2017attention} language model. However, unlike all the aforementioned approaches, it exploits both timestamp and content time during pre-training on a temporal news collection.
As we demonstrate in our experiments, building such a language model 
is both advantageous and of high utility, especially in temporal information retrieval, question answering over temporal collections, and in other NLP tasks that rely on temporal signals. 


\begin{figure*}[t]
\centering
  \includegraphics[width = 0.8\textwidth, height=25mm]{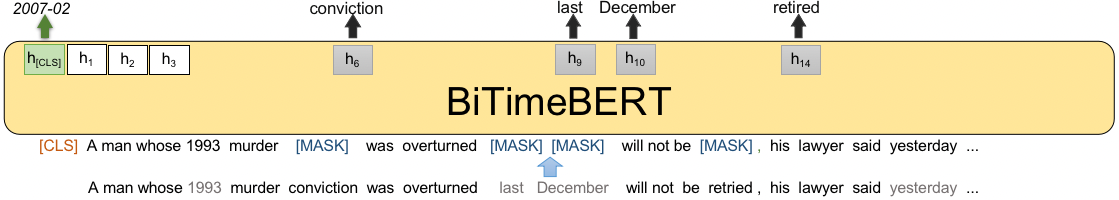}
  \vspace{-1.1em}
  \caption{An illustration of BiTimeBERT training, which includes the TAMLM and DD tasks.}
  \label{fig1}
  \vspace{-1.2em} 
\end{figure*}

\begin{figure}[t]
\centering
  \includegraphics[width = 0.48\textwidth, height=16mm]{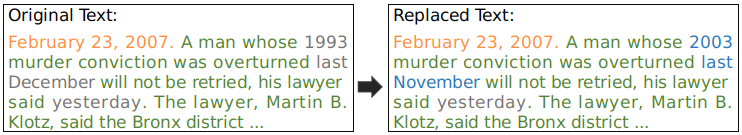}
  \vspace{-2.3em}
  \caption{Example of the replacement procedure in TIR task.} 
  \label{fig2}
  \vspace{-1.3em} 
\end{figure}

\vspace{-0.3em}
\section{Method}
\vspace{-0.2em}
In this section, we present BiTimeBERT, the pre-trained language representation model based on Transformer encoder \cite{vaswani2017attention}. As mentioned before, the model is trained on a temporal collection of news articles via two new pre-training tasks, which involve document timestamp and content time (i.e., the temporal expressions embedded in the content) to construct time-aware language representations. Our approach is inspired by BERT \cite{devlin2018bert}, but distinguishes itself from it in three ways.
Firstly, it is trained on a news article collection spanning two decades rather than on synchronic datasets (such as Wikipedia or Web crawl). 
Note that even if some language models use news article datasets for training (e.g., RoBERTa \cite{liu2019roberta} uses CC-NEWS \cite{nagel2016cc}), they still utilize the same training technique as on the synchronic document collections, which essentially ignores the temporal aspects of documents. Secondly, we use a different masking scheme, \emph{time-aware masked language modeling} (TAMLM) to randomly mask spans of temporal information first rather than just randomly sample tokens from the input. This explicitly forces the model to incorporate both domain knowledge of news archive and temporal information embedded in the document content. Finally, we replace the next sentence prediction (NSP) with an auxiliary objective, \emph{document dating} (DD), which also lets the model incorporate timestamp information while training. As document dating is a type of time prediction, this objective introduces time-related and task-oriented knowledge to the model, and should also aid in improving the performance of other time-related tasks.
Figure~\ref{fig1}\footnote{The selected example is the news article published in The New York Times on 2007/02/23, with the title "Bronx: No Retrial in Murder Case".} illustrates the two proposed objectives. 
BiTimeBERT is jointly trained on the two proposed tasks of TAMLM and DD, with two different additional layers based on the output of its Transformer network.
Moreover, we also propose and test another third pre-training task, \emph{temporal information replacement} (TIR), which, same as TAMLM, makes use of content time, and which, as we found, achieves relatively good performance in some time-related downstream tasks. Figure~\ref{fig2} gives a simple example of the replacement procedure in TIR. All these objectives use cross entropy as the loss function. We describe them in the following sections.

\vspace{-0.4em}
\subsection{Time-aware Masked Language Modeling}
As mentioned above, the first pre-training objective, time-aware masked language modeling (TAMLM), explicitly introduces content time (the temporal information embedded in the document content) during pre-training. This kind of temporal information could be used in understanding the developments of events and identifying the relations between events referred to in text. For example, as discussed earlier, temporal expressions in news (especially ones that refer to past events) have been already used for constructing timeline summaries of temporal news collections \cite{yu2021multi}.

Suppose there is a token sequence $X = (x_{1}, x_{2},..., x_{n})$, where $x_{i}$ $(1\le i \le n)$ indicates a token in the vocabulary. 
First, the temporal expressions in document content are recognized using spaCy (as indicated by the gray font at the bottom in Figure~\ref{fig1}). The recognized temporal expression set is denoted by $T = (t_{1}, t_{2},..., t_{m})$, where $t_{i}$ $(1\le i \le m)$ indicates a particular temporal expression found in the document. Second, unlike in the case of BERT where 15\% of the tokens are randomly sampled in a direct way, we first focus on the extracted temporal expressions. 
30\%\footnote{We chose 30\% as it gives the best performance in most cases after testing models with different percentages in TAMLM task.} of the entire temporal expressions in $T$ are then randomly sampled (e.g., "last December" in Figure~\ref{fig1}). 
Third, we continuously randomly sample other tokens which are not the tokens in $T$, until 15\% of the tokens in total are sampled and masked (e.g., in Figure~\ref{fig1}, "conviction" is masked while "1993" and "yesterday" are not selected to be masked). 
Finally, same as in BERT, 80\% of the sampled tokens are replaced with [MASK], 10\% with random tokens, and 10\% with the original tokens.

Through this masking scheme, we encourage the model to focus on the 
domain knowledge (news article collection in our case) as well as the
content's temporal information. This objective forces the model to incorporate not only the knowledge of the related events, but also the relations between temporal expressions that are not masked when predicting the tokens of masked temporal expressions. For example, in Figure~\ref{fig1}, the masked temporal expression is associated with the overturning of a particular murder conviction that took place in 1993.

\vspace{-0.8em}
\subsection{Document Dating}
\vspace{-0.3em}
The second pre-training objective, document dating (DD), incorporates document timestamp during pre-training. In news archives, each article is usually annotated with a timestamp, corresponding to the date when it was published. As mentioned before, timestamp information can be applied in retrieval, for example, it has been often utilized in temporal information retrieval for estimating document relevance scores \cite{li2003time, kanhabua2010determining, wang2021improving}.

Similar to BERT, the [CLS] token is inserted at the beginning of the input, and its representation, $h_{[CLS]}$, is utilized to provide the contextual representation of the entire token sequence. However, rather than performing binary classification for the next sentence prediction, we utilize this token to predict the document timestamp, as shown in Figure~\ref{fig1}. Temporal granularity of timestamp,\footnote{E.g., the timestamp of an article published in "2007/02/23" under day granularity becomes "2007/02" under month granularity, and "2007" under year granularity.} denoted by $g$, is an important hyper-parameter in this task since timestamp information can be represented at year, month or day temporal granularity. The example shown in Figure~\ref{fig1} uses month granularity.

\citet{jatowt2011extracting} investigate different granularities in news articles showing that time distance and time granularity in news articles are inter-related. \citet{wang2021event} also test their proposed model trained at different granularities for the event time estimation task, and the time is estimated using the same granularity as in the training step.
Thus, the choice of $g$ in BiTimeBERT should also have an effect on the results of downstream tasks.
Loosely speaking, the coarser the granularity, the easier is for the model to predict the timestamp during pre-training, however, the model trained on coarse granularity (e.g., year granularity) might not perform well on difficult time-related tasks. In Section \ref{tempgra_DD}, we analyze the effect of different choices of $g$.


The DD objective incorporates timestamp information during the pre-training phase, which represents the time point at which each document in the pre-training corpus was published. This objective actually introduces task-oriented knowledge to the language model, which strengthens the model on time-related tasks, especially the tasks with a small number of fine-tuning examples - insufficient for training using task-agnostic language models.
Other studies also adapt their language models to the task-specific knowledge via task-oriented pre-training objectives, and show good results after fine-tuning on the corresponding target tasks.
For example, Sentilare model \cite{ke2019sentilare}, which we introduced in Section 2.1, is trained to classify the sentence sentiment during pre-training and then achieves good results on sentiment analysis task. \citet{han2021fine} pre-train their model via MLM together with the proposed utterance relevance classification objective, and afterwards also demonstrate that it performs well on response selection task. Similarly, \citet{xu2019bert} continually train BERT via reading comprehension objective with good results on the review reading comprehension task.


\vspace{-0.3em}
\subsection{Temporal Information Replacement}
We also experiment with one more way in which temporal information of documents could be utilized while pre-training. The last pre-training task we investigate has been inspired by WKLM \cite{xiong2019pretrained}. The authors prove that entity replacement objective can help to capture knowledge about real-world entities. We devise a similar objective called temporal information replacement (TIR) that aims at training the model to capture temporal information of the document content. 
Similar to WKLM that replaces entities of the same type (e.g., the entities of PERSON type can only be replaced with other entities of PERSON type), we enforce the replaced temporal expressions to be of the same temporal granularity. 
First, the timestamp information is inserted at the beginning of the document content and will not be replaced or predicted in the latter steps. This information is useful for the model to understand relative temporal information, e.g., in Figure~\ref{fig2}, "February 23, 2007" could help to infer the actual date denoted by "yesterday". 
We then collect temporal expressions in the news articles using SUTime \cite{chang2012sutime}, a popular tool for recognizing and normalizing temporal expressions, 
and then group those temporal expressions at year, month, and day granularities.\footnote{E.g., "1993" is under year granularity, and an implicit temporal expression like "yesterday" with the corresponding article's timestamp information "2007/02/23" is resolved and converted to "2007/02/22" under day granularity, etc.} Then, 50\% of the time, the temporal expressions of the input sequence are replaced by other temporal expressions, which are randomly sampled from the collected temporal expressions' set of the same granularity, while no replacement is done for the other 50\%. For example, in Figure~\ref{fig2}, "1993" is replaced by "2003" (note that both are of the same granularity), while "yesterday" is not replaced. Then, similar to WKLM, for each temporal expression, the final representations of its boundary words (words before and after the temporal expression) are concatenated and used to make a binary prediction ("replaced" vs. "not replaced").


Note that TIR is an alternative task of TAMLM which also utilizes the content temporal information, yet it is based on swapping instead of masking. However, as will be shown later, our experiments demonstrate that this task can even decrease performance in some downstream tasks. Thus it is not used in the final model of BiTimeBERT. 

\vspace{-0.6em}
\section{Experimental Settings}
\subsection{Pre-training Dataset and Implementation}
For the experiments, we use the New York Times Annotated Corpus (NYT corpus) \cite{sandhaus2008new} as the underlying dataset for pre-training. The NYT corpus contains over 1.8 million news articles published between January 1987 and June 2007, and has been frequently used in Temporal Information Retrieval researches \cite{campos2015survey, DBLP:journals/ftir/KanhabuaBN15}. 
Note that before the pre-training, we randomly sample and remove 50,000 articles from the NYT corpus to use them for running experiments on the document dating downstream task (introduced in Section 4.2), thus these articles are excluded from our pre-training dataset.

As our method can adapt to all the Transformer encoder-based language models, we use BERT \cite{devlin2018bert} as the base framework. Considering the high cost of training from scratch, we utilized the parameters of pre-trained $BERT_{BASE}$ (cased) to initialize our model. BiTimeBERT was continually pre-trained on the NYT corpus for 10 epochs with the TAMLM and the DD task.\footnote{The experiments took about 80 hours on 1 NVIDIA A100 GPU.} The maximum sequence length was 512, while the batch size was 8. We used AdamW \cite{kingma2014adam} as the optimizer and set the learning rate to be 3e-5, with gradient accumulation equal to 8. Finally, the monthly temporal granularity was used in DD task.\footnote{We will study the effect of temporal granularity in DD task in Section \ref{tempgra_DD}.}


\vspace{-0.4em}
\subsection{Downstream Tasks}
We first test our proposal on four datasets of two time-related downstream tasks. These tasks require predicting 
\emph{event occurrence time} (EventTime dataset \cite{wang2021event} and WOTD dataset \cite{honovich2020machine}) and \emph{document timestamp} (NYT-Timestamp dataset and TDA-Timestamp dataset). Note that as current time-aware language models (e.g., TempoBERT \cite{rosin2022time}) have not been originally tested on these two tasks, we do not discuss them in this section. However, in Section 5.5, we evaluate the performance of BiTimeBERT on three datasets of two other tasks that other time-aware language models have been tested in the past (i.e., semantic change detection and sentence time prediction).

The details of 4 datasets we first use are discussed below:

\begin{table}
\footnotesize
  \caption{Sample data from our datasets of time-related tasks.}
  \vspace{-1.5em}
  \label{tab_examples}
  \renewcommand{\arraystretch}{1.3}
  \setlength{\tabcolsep}{0.09em}
  \begin{tabular}{|P{1.13cm}|P{6.6cm}|P{1.1cm}|}
  \hline
  \makecell{\textbf{Dataset}} & \textbf{Text (Event Description or Document Content)} & \makecell{\textbf{Time}}\\
    \hline
   EventTime  & \makecell[l]{Nineteen European nations agree to forbid human cloning.} &  1998-01-12\\
    \hline
    WOTD  & \makecell[l]{American Revolution: British troops occupy Philadelphia.} &  1777\\
    \hline
    \makecell[c]{NYT- \\Timestamp}  & \makecell[l]{It was a message of support and encouragement that Secretary \\of State Warren Christopher delivered to President Boris ...} &  1989-10-09\\
    \hline
     \makecell[c]{TDA- \\ Timestamp}   & \makecell[l]{The Comnaissioners appointed to inquire into the alleged corrupt\\pratctices at Norwich havo made, their report. It cnmmences ...} &  1876-03-20\\
    \hline
\end{tabular}
\vspace{-1.7em} 
\end{table}

\begin{table}
\footnotesize
  \caption{Statistics of the datasets.}
  \vspace{-1.5em}
  \label{tab_datasets}
  \renewcommand{\arraystretch}{1.0}
  \setlength{\tabcolsep}{0.09em}
   \begin{tabular}{|P{1.15cm}|P{0.73cm}|P{1.2cm}|P{2.1cm}|P{1.4cm}|P{1.9cm}|}
  \hline
  \makecell{\textbf{Dataset}} & \textbf{Size} & \textbf{Time Span} & \textbf{Source} &\textbf{Granularity} & \textbf{Task} \\
    \hline
   EventTime  & 22,398 &  1987-2007&   \makecell[c]{Wikipedia \& "On \\This Day" Website} & \makecell[c]{Day, Month,\\ Year} & \makecell[c]{Event \\ Time Estimation}\\
    \hline
    WOTD  & 6,809 &  1302-2018 &  Wikipedia Website & Year & \makecell[c]{Event \\ Time Estimation} \\
    \hline
    \makecell[c]{NYT- \\Timestamp}  & 50,000& 1987-2007& 
    News Archive & \makecell[c]{Day, Month,\\ Year} & \makecell[c]{Document Dating}\\
    \hline
     \makecell[c]{TDA- \\ Timestamp}  & 50,000  & 1785-2009& News Archive &   \makecell[c]{Day, Month,\\ Year} & \makecell[c]{Document Dating}\\
     
    \hhline{|=|=|=|=|=|=|}
    \makecell[c]{\emph{NYT-} \\ \emph{Corpus}}  & \makecell[c]{\emph{1.8} \\ \emph{Million}}& \emph{1987-2007}&  \emph{News Archive}&   \makecell[c]{\emph{Day, Month,}\\ \emph{Year}} & \makecell[c]{\emph{Pre-training}}\\
    
    \hline
\end{tabular}
\vspace{-1.6em} 
\end{table}

\begin{enumerate}[wide, labelwidth=!, labelindent=3pt]
\item \textbf{EventTime} \cite{wang2021event}: This dataset consists of the descriptions and occurrence times of 22,398 events (between January 1987 and June 2007) that were originally collected from Wikipedia year pages\footnote{\url{https://en.wikipedia.org/wiki/List_of_years}} and "On This Day" website.\footnote{\url{https://www.onthisday.com/dates-by-year.php}} 
We will compare our approach with the SOTA method for this dataset. As the SOTA method \cite{wang2021event} conducts search on the entire NYT corpus
, we create an additional dataset called EventTime-WithTop1Doc, with the objective to simulate a similar input setting as in \cite{wang2021event}. The top-1 relevant document of each event in the NYT corpus is firstly extracted using the same retrieval method (BM25) as in \cite{wang2021event}, and the new model input is provided containing the target event description together with appended timestamp and text content of the top-1 document. 
\vspace{0.3em}
\item \textbf{WOTD} \cite{honovich2020machine}: This dataset was scraped from Wikipedia’s On this day webpages,\footnote{\url{https://en.wikipedia.org/wiki/Wikipedia:On_this_day/Today}, accessed 01/2023.} and 
includes 6,809 short descriptions of events and their occurrence year information. WOTD consists of 635 classes, corresponding to 635 different occurrence years. The earliest year is 1302, while the latest is 2018. The median year is 1855.0 whereas the mean is 1818.7. Moreover, the authors additionally provide several sentences about an event, which they call contextual information (CI).\footnote{For example, the contexual information of the WOTD example in Table~\ref{tab_examples} is "The Loyalists never controlled territory unless the British Army occupied it."}
The contextual information is in the form of relevant sentences extracted from Wikipedia. 
Thus, we test two versions of this dataset, with contextual information (CI) and without it (No\_CI). Note that only year information is given as gold labels, hence the tested models can only predict time at year granularity.
Note also that the time span of WOTD dataset (1302-2018) is much longer (and also older) than the one of the NYT corpus (1987-2007) which we used for pre-training. Hence, we can analyze if the models are robust by using WOTD dataset.
\vspace{0.3em}
\item \textbf{NYT-Timestamp}: To evaluate the models on the document dating task, we use the 50,000 separate news articles of the NYT corpus \cite{sandhaus2008new} as mentioned in Section 4.1. 
\vspace{0.3em}
\item \textbf{TDA-Timestamp:\footnote{\url{https://www.gale.com/binaries/content/assets/gale-us-en/primary-sources/intl-gps/ghn-factsheets-fy18/ghn_factsheet_fy18_website_tda.pdf}}} We also test the document dating task on another news corpus, the Times Digital Archive (TDA). TDA contains over 12 million news articles published across more than 200 years (1785-2012),
\footnote{Note that despite TDA containing more articles and spanning a longer time period, the high number of OCR errors in TDA was the reason why we decided not to use it for pre-training but only for testing. Compared with the NYT, the errors are relatively common in TDA (see for example, the last row in Table~\ref{tab_examples}). \cite{niklas2010unsupervised} shows that TDA has a high OCR error rate, especially, in the early years. The average error rate from 1785 to 1932 was found to be above 30\%, while the highest rate can even reach about 60\%.} 
and the time frame of timestamp information of the 50,000 articles that we randomly sampled from this dataset ranges from "1785/01/10" to "2009-12-31". We think that, similarly to WOTD dataset, such a long time span could also help in comparing the 
robustness of different models. 
\end{enumerate}
Same as \cite{wang2021event} and \cite{honovich2020machine} who use a 80:10:10
split to divide EventTime and WOTD, we also divide the constructed NYT-Timestamp, and TDA-Timestamp using the same ratio. 
Table~\ref{tab_datasets} summarizes the basic statistics of the four datasets for downstream tasks along with  describing also our pre-training corpus (i.e., the NYT corpus), while Table~\ref{tab_examples} presents the examples. As we can see in Table~\ref{tab_datasets}, WOTD and TDA-Timestamp have much longer time spans than the one of the pre-training corpus.
As shown in Table~\ref{tab_examples}, the examples of EventTime, NYT-Timestamp, and TDA-Timestamp consist of either detailed occurrence date information or of timestamp information. Therefore, the models tested on these three datasets can be fine-tuned to estimate the time with different temporal granularities. On the other hand, models fine-tuned on WOTD can only predict the time under a year granularity. 
Naturally, the dataset difficulty increases when the time is estimated at finer granularities (e.g., month or day), as the number of labels will also greatly increase. For example, for TDA under day granularity, the label count 
equals to 29,551 which corresponds to the number of days in the dataset.

Note that as event occurrence time estimation requires predicting the time of a given short event description, it is similar to the temporal query analysis (or temporal query profiling) \cite{jones2007temporal, campos2015survey, kanhabua2016temporal}, which aims to identify the time of the interest of short queries, and plays a significant role in temporal information retrieval so that time of queries and time of documents can be matched. Another example of how event occurrence time can be used in practice is in Question Answering over temporal document collections. 
In this kind of QA, a question that does not contain any temporal expression
can be first mapped to its corresponding time period (i.e., time period when the event underlying the question took place) so that the documents from that period can be then processed by a document reader module \cite{wang2020answering, wang2021improving}.\footnote{In Section 5.3 and Section 5.4 we will actually experiment with BiTimeBERT applied in temporal query profiling and temporal question answering, respectively.} 

\vspace{-0.5em}
\subsection{Evaluation Metrics}
As all the above downstream tasks predict time, we use accuracy (ACC) and mean absolute error (MAE) for evaluation, same as \cite{wang2021event}. 

\vspace{-0.4em} %
\begin{enumerate}[label=\arabic*),leftmargin=1.8em]
\item \textbf{Accuracy (ACC)}: The percentage of the events whose occurrence time is correctly predicted.
\item \textbf{Mean absolute error (MAE)}: The average of the absolute differences between the predicted time and the correct occurrence time, based on the specified granularity.
\end{enumerate}

\vspace{-0.4em}
Note that except WOTD dataset, which contains only year information, all models could be evaluated under all the three temporal granularities (i.e., day, month and year). However, as all the pre-trained language models achieve rather poor results under day granularity,\footnote{Still BiTimeBERT outperforms other language models (e.g., under day granularity of EventTime-WithTop1Doc, the ACC score of BiTimeBERT is 2.07, while the scores of BERT, BERT-NYT and BERT-TIR are only 0.04, 0.13 and 0.09, respectively.)} 
we decided to report the results for all granularities only when analyzing the effect of different choices of granularities in DD task (Section \ref{tempgra_DD}).
In particular, in Section \ref{tempgra_DD} we aim to investigate whether the performance of BiTimeBERT could be improved when using a day granularity in DD task.

\vspace{-0.5em}
\subsection{Tested Models}
\vspace{-0.2em}

We test the following models:
\vspace{-0.3em}
\begin{enumerate}[wide, labelwidth=!, labelindent=3pt]
\item \textbf{RG}: Random Guess. The results are estimated by random guess, and the average of 1,000 random selections is used.
\item \textbf{BERT}: The $BERT_{BASE}$ (cased) model 
\cite{devlin2018bert}.
\item \textbf{BERT-NYT}: The $BERT_{BASE}$ (cased) that is continually pre-trained on the NYT corpus for 10 epochs with MLM and NSP tasks.
\item \textbf{SOTA}: SOTA results of EventTime and WOTD, which are taken from \cite{wang2021event} and \cite{honovich2020machine}, respectively. Note that the two methods are not based on language models, and both consist of complex rules or steps of searching and filtering results to obtain the features for estimating the correct date, thus they cannot be easily and quickly applied in other similar tasks.
\item \textbf{BERT-TIR}: The $BERT_{BASE}$ (cased) model continually pre-trained on the NYT corpus for 10 epochs using MLM and TIR.
\item \textbf{BiTimeBERT}: The BiTimeBERT model continually pre-trained on the NYT corpus for 10 epochs using TAMLM and DD tasks.
\end{enumerate}


\vspace{-1.0em}
\subsection{Fine-tuning Setting}
We fine-tune the above language models to the downstream tasks of the four datasets. For each language model, we take the final hidden state of the first token, $h_{[CLS]}$, as the representation of the whole sequence and we add a softmax classifier whose parameter matrix is $X\in \mathbb{R}^{K x H}$, where K is the number of categories of the corresponding dataset. 
In all the settings, we apply a dropout of 0.1 and optimize cross entropy loss using Adam optimizer, with the learning rate equal to 2e-05 and batch size of 16. The maximum sequence length of the models' fine-tuning on EventTime and WOTD is set to 128 as each input is a short event description, while the maximum length on EventTime-WithTop1Doc, NYT-Timestamp, TDA-Timestamp is 512, as their input sequence could be very long.

\vspace{-0.3em}
\section{Experimental Results}
\vspace{-0.2em}
\subsection{Main Results} 
\vspace{-0.2em}
\subsubsection{Event Occurrence Time Estimation} 
Table~\ref{tab_main1} and Table~\ref{tab_main2} present the results of the models on estimating the event occurrence time using EventTime and WOTD, respectively. We first note that BiTimeBERT outperforms other language models,\footnote{The SOTA methods \cite{wang2021event} and \cite{honovich2020machine} are not based on language models.} in ACC and MAE on two datasets over different settings (i.e., year/month granularities, and with/without top1 document information, and with/without contextual information). In addition, we argue that the task is not easy as RG results indicate very poor performance on both datasets.

When considering the year and month granularities of the original EventTime dataset, the improvement comparing BiTimeBERT with BERT is in the range of 47.39\% to 155.21\%, and from 10.09\% to 20.59\% on ACC and MAE metrics, respectively. BiTimeBERT also performs much better than BERT-NYT under both granularities, which achieves similar results as BERT. Moreover, BERT-TIR, the model trained using MLM and the proposed TIR task, shows relatively good performance, too; for example, when comparing with BERT-NYT at year granularity using ACC and MAE, the improvement is 19.53\%, 9.27\%, respectively.

When considering EventTime-WithTop1Doc dataset in which the top-1 document is taken into account for the language models, a significant improvement of BiTimeBERT can be observed. For example, at month granularity using ACC and MAE, the improvement is 98.31\% and 17.05\%, respectively. In addition, BiTimeBERT outperforms BERT by an even larger margin at month granularity, with the improvement of 330.77\%, 23.95\% on ACC and MAE, respectively. BERT-TIR also surpasses BERT with the improvement of 184.45\%, 16.42\%. When comparing with SOTA \cite{wang2021event}, BiTimeBERT achieves similar or even better results under both granularities. 
Moreover, we note that SOTA \cite{wang2021event} requires rather considerable time to prepare the input for time estimation.
Their proposed model utilizes the multivariate time series as the input, which are constructed by analyzing the temporal information of the top-50 retrieved documents and filtering out irrelevant information through several complex steps like sentence similarity computation. 
As we simply use top-1 document ranked by BM25, we believe that the performance of BiTimeBERT could be even further improved by combining with more useful information via more advanced IR techniques.

When considering WOTD dataset, BiTimeBERT outperforms SOTA \cite{honovich2020machine} using accuracy as an evaluation metric, as shown in Table~\ref{tab_main2}. Especially when the contextual information\footnote{As explained in Section 4.2 contextual information contains the relevant sentences extracted from Wikipedia as the external knowledge.} is provided, the improvement is 75.95\%. We also observe that BERT-NYT and BERT-TIR can surpass SOTA \cite{honovich2020machine} and BERT when using contextual information. Note that the two latter methods do not utilize news archives, which suggests that the news archives might be more effective to be used in such a task rather than synchronic document collections (e.g., Wikipedia). 
As BiTimeBERT achieves good performance on WOTD, which is a challenging dataset due to having time span much longer than the one of the pre-training corpus, we think that it has good generalization ability.

\begin{table*}
\begin{minipage}{\columnwidth}
\begin{center}
\footnotesize
 \centering 
 \caption{Performance of different models on EventTime datasets with two different settings.}
 \vspace{-1.7em}
  \label{tab_main1}
  \renewcommand{\arraystretch}{1.0}
  \setlength{\tabcolsep}{0.5em}
  \begin{tabular}{|p{1.2cm}|P{0.56cm}|P{0.56cm}|P{0.56cm}|P{0.56cm}|P{0.56cm}|P{0.56cm}|P{0.56cm}|P{0.56cm}|}
    \hline
    \multirow{3}{3cm}{\textbf{Model}} & \multicolumn{4}{c|}{\textbf{EventTime}}& \multicolumn{4}{c|}{\textbf{EventTime-WithTop1Doc}}\\
    \cline{2-9}
    & \multicolumn{2}{c|}{\textbf{Year}} & \multicolumn{2}{c|}{\textbf{Month}}& \multicolumn{2}{c|}{\textbf{Year}} & \multicolumn{2}{c|}{\textbf{Month}}\\
    \cline{2-9}
    & \textbf{ACC} & \textbf{MAE} & \textbf{ACC} & \textbf{MAE} & \textbf{ACC} & \textbf{MAE}& \textbf{ACC} & \textbf{MAE}\\
    \hline
    RG & 4.77 & 6.92 & 0.41 & 81.60 & 4.77 & 6.92 & 0.40 & 81.70 \\ \hline
    BERT & 21.65 & 3.47 & 5.09 & 43.81 & 35.98 & 3.89 & 5.98 & 37.95 \\ \hline
    BERT-NYT & 21.25 & 3.56 & 5.18 & 43.50 & 34.46 & 4.45 & 8.21 & 34.14 \\ \hline
    SOTA \cite{wang2021event}& - & - & - & - & 40.93 & 3.01 & \textbf{30.89} & 36.19 \\ \hline
    BERT-TIR & 25.40 & 3.23 & 6.83 & 40.45 & 36.47 & 3.54 & 17.01 & 31.72 \\ \hline
    BiTimeBERT &  \textbf{31.91} & \textbf{3.12}& \textbf{12.99} & \textbf{34.79} & \textbf{41.96} & \textbf{2.40}& 25.76 & \textbf{28.86}
    \\ \hline
  \end{tabular}
\end{center}
\vspace{-1.0em}
\end{minipage}\hfill 
\begin{minipage}{\columnwidth}
\begin{center}
\footnotesize
 \centering
  \caption{Performance of different models on WOTD dataset with/without contextual information.}
 \vspace{-1.7em}
  \label{tab_main2}
  \renewcommand{\arraystretch}{1.0}
  \setlength{\tabcolsep}{0.5em}
  \begin{tabular}{|p{1.2cm}|P{0.56cm}|P{0.56cm}|P{0.56cm}|P{0.56cm}|}
    \hline
    \multirow{2}{3cm}{\textbf{Model}} & \multicolumn{2}{c|}{\textbf{NO\_CI}} & \multicolumn{2}{c|}{\textbf{CI}}\\
    \cline{2-5}
    & \textbf{ACC} & \textbf{MAE} & \textbf{ACC} & \textbf{MAE}\\
    \hline
    RG & 0.16 & 217.72 & 0.15 & 217.57\\ \hline
    BERT & 7.20 & 52.58 & 9.69 & 41.16\\ \hline
    BERT-NYT & 8.08 & 53.75 & 19.97 & 36.47\\ \hline
    SOTA \cite{honovich2020machine}& 11.40 & - & 13.10 & - \\ \hline
    BERT-TIR & 10.13 & 54.92 & 18.36 & 35.99\\ \hline
    BiTimeBERT &  \textbf{11.60} & \textbf{48.51} & \textbf{23.05} & \textbf{33.70} 
    \\ \hline
  \end{tabular}
\end{center}
\end{minipage}
\vspace{-2.2em} 
\end{table*}

\begin{table*}
\begin{minipage}{\columnwidth}
\begin{center}
\footnotesize
 \centering 
 \caption{Performance of different models for document dating on NYT-Timestamp and TDA-Timestamp.}
 \vspace{-1.7em}
  \label{tab_main3}
  \renewcommand{\arraystretch}{1.0}
  \setlength{\tabcolsep}{0.5em}
  \begin{tabular}{|p{1.2cm}|P{0.56cm}|P{0.56cm}|P{0.56cm}|P{0.56cm}|P{0.56cm}|P{0.56cm}|P{0.56cm}|P{0.66cm}|}
    \hline
    \multirow{3}{3cm}{\textbf{Model}} & \multicolumn{4}{c|}{\textbf{NYT-Timestamp}}& \multicolumn{4}{c|}{\textbf{TDA-Timestamp}}\\
    \cline{2-9}
    & \multicolumn{2}{c|}{\textbf{Year}} & \multicolumn{2}{c|}{\textbf{Month}}& \multicolumn{2}{c|}{\textbf{Year}}& \multicolumn{2}{c|}{\textbf{Month}}\\
    \cline{2-9}
    & \textbf{ACC} & \textbf{MAE} & \textbf{ACC} & \textbf{MAE} & \textbf{ACC} & \textbf{MAE}& \textbf{ACC} & \textbf{MAE}\\
    \hline
    RG  & 4.77 & 7.06 & 0.41 &81.79 & 0.45 & 75.39 & 0.04 & 873.88 \\ \hline
    BERT & 35.00 & 1.64 & 2.56 & 22.74 &15.84 & 44.87 & 0.80 & 632.66 \\ \hline
    BERT-NYT & 38.74 & 1.41 & 8.24 & 18.35 & 15.04 & 45.16 & 0.66 & 669.02\\ \hline
    BERT-TIR & 48.06 & 1.09 & 20.30 & 13.54 & 17.72 & 43.53 & 1.26 & 589.69\\ \hline
    BiTimeBERT &  \textbf{58.72} & \textbf{0.80}& \textbf{31.10} & \textbf{9.54} & \textbf{19.00} & \textbf{40.11}& \textbf{2.38} & \textbf{580.25} 
    \\ \hline
  \end{tabular}
\end{center}
\vspace{-1.6em}
\end{minipage}\hfill 
\begin{minipage}{\columnwidth}
\begin{figure}[H]
  \includegraphics[width = \textwidth, height=31mm]{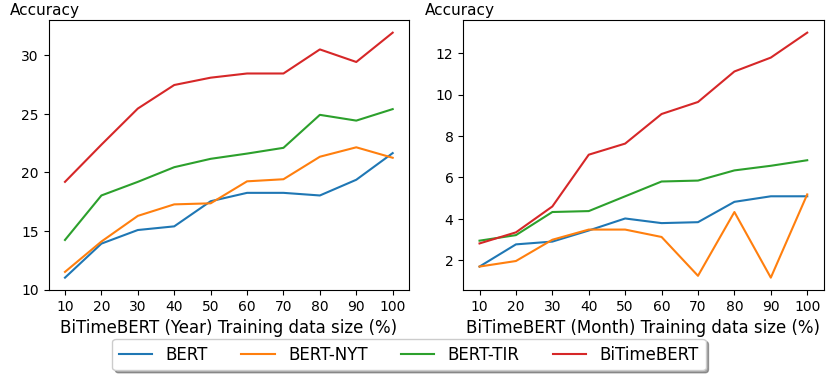}
  \vspace{-2.6em}
  \caption{Impact of training data size (best viewed in color).}
  \label{data_size}
\end{figure}
\end{minipage}
\vspace{-1.1em} 
\end{table*}

\vspace{-0.5em}
\subsubsection{Document Dating} 
Table~\ref{tab_main3} presents the results of the document dating tasks. 
All the language models achieve weak results under month granularity at TDA-Timestamp, likely due to TDA-Timestamp having 2,627 month labels. In addition, the timestamps in the 50,000 articles of TDA-Timestamp range from 1785 to 2009, which further increases the difficulty. 
We thus mainly compare the models on NYT-Timestamp of year and month granularities, and on TDA-Timestamp of year granularity. BiTimeBERT still outperforms other language models with substantial gains. When considering the year and month granularities of NYT-Timestamp, the improvement comparing BiTimeBERT with BERT-NYT is in the range of 51.57\% to 277.43\%, and from 43.26\% to 48.01\% on ACC and MAE metrics, respectively. When considering TDA-Timestamp under year granularity, the improvement is 26.33\% and 11.18\% on ACC and MAE, respectively. In addition, BERT-TIR also obtains relatively good results on both datasets, suggesting that the TIR task is also effective, however substantially less than when using BiTimeBERT.

\vspace{-0.5em} 
\subsection{Additional Analysis} 
\subsubsection{Ablation Study} 
\label{Ablation}

\begin{table*}
\begin{minipage}{\columnwidth}
\begin{center}
\footnotesize
 \centering 
 \caption{Ablation test on event occurrence time estimation.}
 \vspace{-1.9em}
  \label{tab_abl1}
  \renewcommand{\arraystretch}{1.0}
  \setlength{\tabcolsep}{0.2em}
  \begin{tabular}{|p{1.3cm}|P{0.66cm}|P{0.8cm}|P{0.66cm}|P{0.8cm}|P{0.66cm}|P{0.8cm}|P{0.66cm}|P{0.8cm}|}
    \hline
    \multirow{3}{3cm}{\textbf{Model}} & \multicolumn{4}{c|}{\textbf{EventTime}}& \multicolumn{4}{c|}{\textbf{WOTD}}\\
    \cline{2-9}
    & \multicolumn{2}{c|}{\textbf{Year}} & \multicolumn{2}{c|}{\textbf{Month}}& \multicolumn{2}{c|}{\textbf{NO\_CI}}& \multicolumn{2}{c|}{\textbf{CI}}\\
    \cline{2-9}
    & \textbf{ACC} & \textbf{MAE} & \textbf{ACC} & \textbf{MAE} & \textbf{ACC} & \textbf{MAE}& \textbf{ACC} & \textbf{MAE} \\
    \hline
    TAMLM & 23.05 & 3.37 & 6.87 & 41.16 & 9.43 & 53.48 & 19.82 & 38.74\\ \hline
    DD  & 24.81 & 3.41 & 7.02 & 41.62 & 9.56 & 60.14 & 18.42 & 40.64\\ \hline
    MLM  & 21.52 & 3.45 & 5.71  & 44.47 & 8.66 & 55.66 & 18.80 & 40.85\\ \hline
    MLM+DD & 25.05 & 3.63 & 7.92 & 40.36 & 10.51 & 59.74 & 19.12 & 42.14\\ \hline
    BiTimeBERT &  \textbf{29.51} & \textbf{3.17}& \textbf{10.80} & \textbf{36.11}& \textbf{11.16} & \textbf{51.09}& \textbf{22.47} & \textbf{36.80}
    \\ \hline
  \end{tabular}
\end{center}
\end{minipage}\hfill 
\begin{minipage}{\columnwidth}
\begin{center}
\footnotesize
 \centering 
 \caption{Ablation test on document dating.}
 \vspace{-1.8em}
  \label{tab_abl2}
  \renewcommand{\arraystretch}{1.0}
  \setlength{\tabcolsep}{0.4em}
  \begin{tabular}{|p{1.2cm}|P{0.66cm}|P{0.66cm}|P{0.66cm}|P{0.66cm}|P{0.66cm}|P{0.66cm}|P{0.66cm}|P{0.8cm}|}
    \hline
    \multirow{3}{3cm}{\textbf{Model}} & \multicolumn{4}{c|}{\textbf{NYT-Timestamp}}& \multicolumn{4}{c|}{\textbf{TDA-Timestamp}}\\
    \cline{2-9}
    & \multicolumn{2}{c|}{\textbf{Year}} & \multicolumn{2}{c|}{\textbf{Month}}& \multicolumn{2}{c|}{\textbf{Year}}& \multicolumn{2}{c|}{\textbf{Month}}\\
    \cline{2-9}
    & \textbf{ACC} & \textbf{MAE} & \textbf{ACC} & \textbf{MAE} & \textbf{ACC} & \textbf{MAE} & \textbf{ACC} & \textbf{MAE} \\
    \hline
    TAMLM & 39.92 & 1.46 & 8.80 & 16.74 & 14.96 & 45.80 & 0.95 & 623.14\\ \hline
    DD  & 49.86 & 1.32 & 21.74 &14.05 & 15.61 & 46.23 & 1.23 & 622.25  \\ \hline
    MLM  & 36.98 & 1.51 & 3.46 & 19.17 & 14.44 & 46.08 & 0.64 & 693.14 \\ \hline
    MLM+DD & 51.48 & 1.19 & 23.86 & 14.18 & 15.34 & 45.91 & 1.24 & 616.51\\ \hline
    BiTimeBERT &  \textbf{56.08} & \textbf{0.81}& \textbf{27.42} & \textbf{10.56} & \textbf{18.54} & \textbf{43.00} & \textbf{1.94} & \textbf{595.47}
    \\ \hline
  \end{tabular}
\end{center}
\end{minipage}
\vspace{-1.3em} 
\end{table*}

To study the effect of the two objectives of BiTimeBERT, we next conduct an ablation analysis and present its results in Table~\ref{tab_abl1} and Table~\ref{tab_abl2}. We compare in total five models that use different pre-training tasks and test them on the four datasets. 
\textbf{DD}, \textbf{TAMLM}, \textbf{MLM} indicate the corresponding models trained using only DD, TAMLM or MLM tasks, respectively. \textbf{MLM+DD} means the model is trained using both BERT's MLM task and our proposed DD objective. For fair and effective comparison, all five models continually pre-train $BERT_{BASE}$ with their specific pre-training tasks on the NYT corpus for 3 epochs.

As shown in Table~\ref{tab_abl1} and Table~\ref{tab_abl2}, BiTimeBERT, which uses TAMLM and DD as the pre-training tasks, achieves the best results across all the datasets, suggesting that the two proposed objectives contribute to the performance of our model. When considering the models that use only one of the pre-training objectives of BiTimeBERT, TAMLM or DD, the performance is better than MLM in most cases. This confirms that the two proposed pre-training tasks of BiTimeBERT are both helpful in obtaining effective time-aware language representations of text. 
Yet, incorporating at the same time the two proposed objectives of BiTimeBERT that make use of different temporal aspects produces the best results.

\vspace{-0.6em} 
\subsubsection{Effect of Different Temporal Granularities in DD}
\label{tempgra_DD}
We examine now BiTimeBERT training using different settings for the temporal granularity $g$ in DD objective. We first pre-train different BiTimeBERT variants with three different $g$ for 3 epochs, and then fine-tune the models on four datasets. The models of different granularities are denoted by \textbf{BiTimeBERT-Year}, \textbf{BiTimeBERT-Month} and \textbf{BiTimeBERT-Day}. As shown in Table~\ref{tab_temp1}, BiTimeBERT-Month achieves the best results most of the time, while BiTimeBERT-Day performs poorly in some "easy" tests, e.g., for the EventTime and NYT-Timestamp of year granularity, as well as WOTD with CI.
We also observe that none of the models can produce relatively good performance on the hard tasks (e.g., EventTime of day granularity). This might be mainly due to: (1) the models are still under-fitting and may need to be trained with more epochs, especially, at day granularity in DD task, and (2) more data is needed for pre-training. 

\begin{table*}[]
\begin{center}
\footnotesize
 \centering \caption{BiTimeBERT with different temporal granularities on event occurrence time estimation and document dating.}
 \vspace{-1.8em}
  \label{tab_temp1}
  \renewcommand{\arraystretch}{1.0}
  \setlength{\tabcolsep}{0.3em}
  \begin{tabular}{|p{2.1cm}|P{0.5cm}|P{0.5cm}|P{0.5cm}|P{0.5cm}|P{0.5cm}|P{0.8cm}|P{0.5cm}|P{0.5cm}|P{0.5cm}|P{0.5cm}|P{0.5cm}|P{0.5cm}|P{0.5cm}|P{0.5cm}|P{0.45cm}|P{0.65cm}|P{0.5cm}|P{0.5cm}|P{0.45cm}|P{0.65cm}|P{0.45cm}|P{0.9cm}|}
    \hline
    \multirow{3}{3cm}{\textbf{Model}} & \multicolumn{6}{c|}{\textbf{EventTime}}& \multicolumn{4}{c|}{\textbf{WOTD}}&
    \multicolumn{6}{c|}{\textbf{NYT-Timestamp}}& \multicolumn{6}{c|}{\textbf{TDA-Timestamp}}\\
    \cline{2-23}
    & \multicolumn{2}{c|}{\textbf{Year}} & \multicolumn{2}{c|}{\textbf{Month}}& \multicolumn{2}{c|}{\textbf{Day}} & \multicolumn{2}{c|}{\textbf{NO\_CI}}& \multicolumn{2}{c|}{\textbf{CI}}&
    \multicolumn{2}{c|}{\textbf{Year}} & \multicolumn{2}{c|}{\textbf{Month}}& \multicolumn{2}{c|}{\textbf{Day}} & \multicolumn{2}{c|}{\textbf{Year}}& \multicolumn{2}{c|}{\textbf{Month}}& \multicolumn{2}{c|}{\textbf{Day}}\\
    
    \cline{2-23}
    & \textbf{ACC} & \textbf{MAE} & \textbf{ACC} & \textbf{MAE} & \textbf{ACC} & \textbf{MAE}& \textbf{ACC} & \textbf{MAE} & \textbf{ACC} & \textbf{MAE}& \textbf{ACC} & \textbf{MAE} & \textbf{ACC} & \textbf{MAE} & \textbf{ACC} & \textbf{MAE}& \textbf{ACC} & \textbf{MAE} & \textbf{ACC} & \textbf{MAE}& \textbf{ACC} & \textbf{MAE}\\
    \hline
    BiTimeBERT-Year  & \textbf{30.71} & \textbf{3.06} & 8.62 & 38.35 & 0.76 & 1772.48 & 9.84 & 59.76 & 20.56 & \textbf{35.67} & \textbf{57.48} & \textbf{0.78} & 19.46 & 11.30 & 0.34 & 401.88 & 17.88 & 43.93 & 1.02 & \textbf{575.04} & 0.00 & 14168.61\\ \hline
    BiTimeBERT-Month & 29.51 & 3.17 & \textbf{10.80} & \textbf{36.11} & \textbf{1.83} & 1743.75 & \textbf{11.16} & \textbf{51.09} & \textbf{22.47} & 32.92 & 56.08 & 0.81 & \textbf{27.42} & \textbf{10.56} & \textbf{0.72} & 406.52 & \textbf{18.54} & \textbf{43.00}  & \textbf{1.30} & 643.38& \textbf{0.02} & \textbf{12083.72}\\ \hline
    BiTimeBERT-Day  & 26.43 & 3.18 & 7.99 & 38.42 & 1.27 & \textbf{1647.64} & 10.72 & 53.36 & 17.47 & 40.22  & 54.06 & 0.91 & 19.46 & 11.02 & 0.64 & \textbf{398.77} & 18.08 & 43.41 & 1.14 & 603.71 & 0.00 & 13794.74\\ \hline
  \end{tabular}
\end{center}
\vspace{-1.8em} 
\end{table*}

\vspace{-0.6em} 
\subsubsection{Data Size Analysis} Figure~\ref{data_size} (left) and Figure~\ref{data_size} (right) plot the accuracy of four pre-trained language models on EventTime of various sizes of training data, under year and month granularities, respectively. First of all, BiTimeBERT consistently performs better than other models using the same size of training data, and can achieve the similar best performance of other models by using much less data. In addition, especially under month granularity, we can observe a clear increasing trend of the accuracy of BiTimeBERT model. 
It might be even able to
achieve new SOTA performance if more data is used, while BERT and BERT-TIR models exhibit less performance gain when using more data.

\vspace{-0.6em} 
\subsubsection{Temporal Semantic Similarity Analysis} 
\label{similarity_section} 
We now perform simple similarity experiments without fine-tuning in order to measure whether BiTimeBERT indeed generates effective time-aware language representations when it is not adapted to any particular downstream task. 
The EventTime dataset which contains information of events that occurred between January 1987 and June 2007 is used here again. 
In particular, we first collect contextual representations (i.e., the final hidden state vector of [CLS] output by the model) of all possible atomic time units from the range of January 1987 to June 2007, under year and month granularities. 
For example, under year granularity, such a set contains 21 vectors corresponding to the representations of temporal expressions from "1987" to "2007". 
For a given event description, we then compute the cosine similarities between its contextual representation and the representation of each temporal expression in the set. The temporal expression with the largest similarity score is finally considered as the estimated event time for the event.
As shown in Table~\ref{temp_semantic_sim}, BiTimeBERT outperforms BERT and BERT-NYT by a large margin under both granularities, demonstrating that it can construct more effective time-aware language representations, and learns both domain knowledge and task-oriented knowledge even without fine-tuning.

\begin{table}[!]
\begin{center}
\footnotesize
 \centering \caption{Temporal semantic similarity on the EventTime dataset. The models are tested without fine-tuning.}
 \vspace{-1.8em}
  \label{temp_semantic_sim}
  \renewcommand{\arraystretch}{1.0}
  \setlength{\tabcolsep}{0.5em}
  \begin{tabular}{|p{1.2cm}|P{0.66cm}|P{0.8cm}|P{0.66cm}|P{0.8cm}|}
    \hline
    \multirow{2}{3cm}{\textbf{Model}} & \multicolumn{2}{c|}{\textbf{Year}} & \multicolumn{2}{c|}{\textbf{Month}}\\
    \cline{2-5}
    & \textbf{ACC} & \textbf{MAE} & \textbf{ACC} & \textbf{MAE}\\
    \hline
    BERT & 3.03 & 10.47 & 0.13 & 76.97\\ \hline
    BERT-NYT & 4.82 & 7.36 & 0.66 & 76.36\\ \hline
    BERT-TIR & 11.29 & 5.99 & 1.91 & 82.04\\ \hline
    BiTimeBERT &  \textbf{14.33} & \textbf{5.72} & \textbf{3.83} & \textbf{66.35} 
    \\ \hline
  \end{tabular}
\end{center}
\vspace{-1.8em} 
\end{table}

\vspace{-0.6em}
\subsection{Case Study on Time-Sensitive Queries} 

We next conduct two types of small case studies of event time
prediction. This time we apply a challenging setting by using short time-sensitive queries related to events\footnote{Compared with EventTime dataset for which
the average number of tokens of event descriptions is 17.3, the average number of tokens of the queries here is only 3.2.} to estimate their dates under year granularity by applying BiTimeBERT without any fine-tuning. The queries represent non-recurring as well as recurring events. A non-recurring event is an event that occurred at one specific time point (e.g., "9/11 attacks"), while a recurring event is one that occurred multiple times in the past (e.g., "Summer Olympic Games"). Similar to Section \ref{similarity_section}, we compare the cosine similarity of the representations between a query and each temporal expression. 
However, rather than computing ACC and MAE using the date with the largest similarity score, we return a ranked list of dates and calculate MRR (Mean Reciprocal Rank) for the non-recurring event test. This is done in order to find "where is the correct time of non-recurring event located in the list". On the other hand, for the recurring event test, we use MAP (Mean Average Precision) to find "if all occurrence dates of a recurring event are at the top of list".

\vspace{-0.7em} 
\subsubsection{Non-recurring events.} For non-recurring events, we prepared 10 example short queries of September 11 attacks that occurred in 2001: "9/11 attacks", "Aircraft hijackings", "19 terrorists", "Osama bin Laden", "the Twin Towers", "War on terrorism", "American Airlines Flight 77", "American Airlines Flight 11", "United Airlines Flight 175", "United Airlines Flight 93". We then created the ranked list by comparing the similarity between the query vector and vectors of temporal expressions under year granularity, from "1987" to "2007". As shown in Table~\ref{tab_casestudy} (left), BiTimeBERT performs the best, demonstrating that it effectively captures the knowledge of correct temporal information for such event.

\vspace{-0.7em} 
\subsubsection{Recurring events.} For recurring events, we also collected 10 short queries representing important and periodical example events: "Summer Olympic Games", "FIFA World Cup", "Asian Games", "Commonwealth Games", "World Chess Championship", "United States presidential election", "French presidential election", "United Kingdom general election", "United States senate election", "United States midterm election". Note that as these events also occurred before 1987, we additionally compare them with the temporal expressions under year granularity within the time period from "1966" to "1986". Thus, two ranked lists of dates are obtained. 
As shown in Table~\ref{tab_casestudy} (right), BiTimeBERT also performs the best, indicating that most occurrence dates appeared at the top of the list. Moreover, when estimating dates outside the time span of the pre-training corpus, i.e.,  from "1966" to "1986", BiTimeBERT also obtains good performance. 

These results also indicate that BiTimeBERT successfully fuses both domain knowledge and task-oriented knowledge extracted from temporal news collection during the pre-training phase, and is able to construct effective word representations that capture temporal aspects of queries, even very short queries.

\begin{table}
\caption{Results on non-recurring (left) and recurring events (right). The models are tested without fine-tuning.}
\vspace{-0.2em}
\label{tab_casestudy}

\begin{center}
\centering 
\begin{minipage}{0.4\columnwidth}
\begin{center}
\footnotesize
 \centering 
 \vspace{-1.9em}
  \renewcommand{\arraystretch}{1.0}
  \setlength{\tabcolsep}{0.5em}
  \begin{tabular}{|p{1.4cm}|c|}
  \hline
  \multirow{2}{3cm}{\textbf{Model}} &\makecell[c]{\textbf{Non-recurring} \\\textbf{Events}}  \\
  \cline{2-2} & {\textbf{MRR}} \\
    \hline
    \makecell[l]{BERT} & 0.1277 \\
    \hline
    \makecell[l]{BERT-NYT} & 0.3601\\
    \hline
    \makecell[l]{BERT-TIR} & 0.4533 \\
    \hline
    \makecell[l]{BiTimeBERT} & \textbf{0.5417}\\
    \hline
    \end{tabular}
\end{center}

\end{minipage}
\hfill 
\begin{minipage}{0.58\columnwidth}
\begin{center}
\footnotesize
 \centering 
 \vspace{-1.9em}
  \renewcommand{\arraystretch}{1.0}
  \setlength{\tabcolsep}{0.5em}
  \begin{tabular}{|p{1.4cm}|c|c|}
    \hline
    \multirow{3}{3cm}{\textbf{Model}} & \multicolumn{2}{c|}{\textbf{Recurring Events}} \\
    \cline{2-3}
     & \multicolumn{1}{c|}{\textbf{1966-1986}} & \multicolumn{1}{c|}{\textbf{1987-2007}}  \\
    \cline{2-3}
     & \textbf{MAP} & \textbf{MAP}   \\
    \hline
    \makecell[l]{BERT} & 0.4042  & 0.3512\\
    \hline
    \makecell[l]{BERT-NYT} & 0.3633 & 0.3197\\
    \hline
    \makecell[l]{BERT-TIR} & 0.4449 & 0.4661\\
    \hline
    \makecell[l]{BiTimeBERT} & \textbf{0.5294}  & \textbf{0.6686}\\
    \hline
  \end{tabular}
\end{center}
\end{minipage}
\end{center}

\vspace{-1.4em} 
\end{table}

\vspace{-0.6em}
\subsection{Application for Temporal QA} 

\begin{table}
\begin{center}
\footnotesize
  \centering \caption{Performance of different models in QA task.}
  \vspace{-1.8em}
  \label{tab6}
  \renewcommand{\arraystretch}{1.0}
  \setlength{\tabcolsep}{0.6em}
  \centering
  \begin{tabular}{|c|P{0.46cm}|P{0.46cm}|P{0.46cm}|P{0.46cm}|P{0.46cm}|P{0.46cm}|P{0.46cm}|P{0.46cm}|}
    \hline
    \multirow{2}{0.70cm}{\textbf{Model}} & \multicolumn{2}{c|}{\textbf{Top 1}} & \multicolumn{2}{c|}{\textbf{Top 5}}& \multicolumn{2}{c|}{\textbf{Top 10}}& \multicolumn{2}{c|}{\textbf{Top 15}}\\
    \cline{2-9}
    & \textbf{EM} & \textbf{F1} & \textbf{EM} & \textbf{F1} & \textbf{EM} & \textbf{F1} & \textbf{EM} & \textbf{F1}\\
    \hline
    QANA \cite{wang2021improving} & 21.00& 28.90 & 28.20 & 36.85 & 34.20 & 44.01 & 36.20 & 45.63\\ \hline
     QANA+BiTimeBERT &  \textbf{22.40} & \textbf{29.31} & \textbf{29.20} & \textbf{37.14} & \textbf{34.80} & \textbf{44.34} & \textbf{36.40} & \textbf{46.01}
    \\ \hline
  \end{tabular}
  \end{center}
\vspace{-1.5em} 
\end{table}

BiTimeBERT can be used in several ways and supports different applications for which time is important. As we have seen in Sections 5.1.1 and 5.3, it can be easily applied in temporal information retrieval  \cite{alonso2007value,campos2015survey}, for example, aiding in the time-based exploration of textual archives by estimating the time of interest of queries, so that the computed query temporal information could be utilized for time-aware document ranking. Other potential applications are: document dating \cite{jatowt2012large, kanhabua2009using, kotsakos2014burstiness}, 
 temporal image retrieval \cite{dias2012temporal}, 
 event detection and ordering \cite{strotgen2012event,bookdercz}, temporal QA \cite{pasca,wang2020answering}, etc.

We demonstrate here how BiTimeBERT could be utilized in one such application. In particular, we improve a temporal question answering system called QANA \cite{wang2021improving}, which 
achieves good performance in answering event-related questions that are implicitly time-scoped (e.g., "\emph{Which famous painting by Norwegian Edvard Munch was stolen from the National Gallery in Oslo?}" is an implicitly time-scoped question as it does not contain any temporal expression, yet it is implicitly related to temporal information of its corresponding specific event, which is "1994/05"). To answer implicitly time-scoped questions, 
QANA needs to first estimate the time scope of the event described in the question at month granularity, which is then mapped to the time interval with the "start" and "end" information (e.g., one possible time scope of the above-mentioned question example is ("1994/03", "1994/08")). 

Instead of analyzing the temporal distribution of retrieved documents to estimate the time scope as is in QANA's original implementation, we adapt QANA by using the BiTimeBERT fine-tuned on EventTime-WithTop1Doc under month granularity. Similar to the way of constructing EventTime-WithTop1Doc, the top-1 relevant document of each question is first selected using BM25, and then its timestamp and text content are appended to the corresponding questions, which are further sent to BiTimeBERT as an input. We then keep two time points of the top 2 probabilities predicted by BiTimeBERT, which are then ordered and used as "start" and "end" information of the estimated question's time scope. The estimated time scope is then utilized for reranking documents, and finally, the answers are returned by the Document Reader Module of QANA. In other words, in our adaptation of QANA, we only replace the step of the question's time scope estimation. We denote such a modified system as QANA+BiTimeBERT. We test this system on the test set of 500 manually created implicitly time-scoped questions published in \cite{wang2021improving}. 
As the number of the top $N$ re-ranked documents 
affects the final results, we also test the effect of different top $N$ values. As shown in Table~\ref{tab6}, QANA+BiTimeBERT outperforms QANA for all different $N$ values. For example, on top-1 document, the extended model has a 6.67\% improvement on EM and 1.42\% on F1. 




\vspace{-1.0em}
\subsection{ Semantic Change Detection \& Sentence Time Prediction}

\begin{table}
\footnotesize
  \caption{Statistics of semantic change detection datasets.}
  \vspace{-1.8em}
  \label{tab_scd_datasets}
  \renewcommand{\arraystretch}{1.0}
  \setlength{\tabcolsep}{0.09em}
   \begin{tabular}{|P{1.8cm}|P{1.2cm}|P{1.2cm}|P{1.3cm}|P{1.4cm}|P{1.3cm}|}
  \hline
  \makecell{\textbf{Dataset}} & \textbf{\makecell[c]{Target \\ Words}}& \textbf{ C1 Source} & \textbf{\makecell[c]{C1 Time \\ Period}} &\textbf{C2 Source} & \textbf{\makecell[c]{C2 Time \\ Period}} \\
    \hline
   LiverpoolFC  & 97 &  Reddit &   2011–2013 & Reddit & 2017 \\
    \hline
    SemEval-English  & 37 &  CCOHA &   1810–1860 & CCOHA & 1960–2010 \\
    \hline
\end{tabular}
\vspace{-1.8em} 
\end{table}

\begin{table}[!]
\begin{center}
\footnotesize
 \centering \caption{Semantic change detection results.}
 \vspace{-1.8em}
  \label{temp_semantic_change}
  \renewcommand{\arraystretch}{1.0}
  \setlength{\tabcolsep}{0.6em}
  \begin{tabular}{|c|c|c|c|c|}
    \hline
    \multirow{2}{2cm}{\textbf{Model}} & \multicolumn{2}{c|}{\textbf{LiverpoolFC}} & \multicolumn{2}{c|}{\textbf{ SemEval-Eng}}\\
    \cline{2-5}
    & \textbf{Pearson} & \textbf{Spearman} & \textbf{Pearson} & \textbf{Spearman}\\
    \hline
    BERT & 0.414 & 0.454 & 0.483 &  0.416\\ \hline
    BERT-NYT & 0.431 & 0.463 & 0.510 & 0.422\\ \hline
    TempoBERT (cos\_dist) \cite{rosin2022time} & 0.371 & 0.451 & 0.538 & 0.467\\ \hline
    TempoBERT (time-diff)\cite{rosin2022time} & \textbf{0.637} & \textbf{0.620} & 0.208 & 0.381\\ \hline
    BiTimeBERT &  0.468 & 0.476 & \textbf{0.616} & \textbf{0.476} 
    \\ \hline
  \end{tabular}
\end{center}
\vspace{-1.8em} 
\end{table}

\vspace{-0.2em}
In this last section, we compare BiTimeBERT with TempoBERT \cite{rosin2022time}, a recently proposed time-aware BERT model which works by prepending texts with timestamp and then masking the added tokens during training, as discussed in Section 2.2. 
TempoBERT hence does not utilize content time. It has been also tested only on sentence-level corpora which does not assure its generalization ability on other datasets or tasks that have long texts as input, e.g., EventTime-WithTop1Doc dataset or document dating task.
We compare TempoBERT and BiTimeBERT (as well as BERT and BERT-NYT) on the following two time-related tasks which were used by the TempoBERT's authors for evaluating their system \cite{rosin2022time}: 
\vspace{-1.2em}
\begin{enumerate}[wide, labelwidth=!, labelindent=3pt]
\vspace{-0.2em}
\item \textbf{Semantic change detection:} This task requires determining whether and to what extent the meanings of a set of target words have changed over time, with the help of time-annotated corpora. Following TempoBERT, the LiverpoolFC corpus (short-term corpora) \cite{del2018short} and the SemEval-English corpus (long-term corpora) \cite{schlechtweg2020semeval} are used. Table~\ref{tab_scd_datasets} presents the statistics of both datasets. 
To determine how well a model can detect changes in the meaning of words over time, we measure its performance by comparing the model’s assessment of semantic shift for each target word to the semantic index (i.e., the ground truth). The correlation between the two provides a measure of the model's effectiveness in detecting semantic change.
In particular, both Pearson’s correlation coefficient and Spearman’s rank correlation coefficient are calculated. 
For fair comparison, we adopt the same training hyperparameters as TempoBERT: the learning rate and epochs number for LiverpoolFC are 1e-7 and 1, respectively, while they are 1e-6 and 2 for SemEval-English. However, as the corpora of sentence level seldom contain the content temporal information for TAMLM objective, we train BiTimeBERT using MLM for domain adaptation which is also used in training BERT and BERT-NYT. 
After obtaining the trained language models, we apply on them the method used in TempoBERT to generate representations of target words for each time period and to estimate the semantic change of words by measuring cos\_dist (cosine distance). 
Note that \citet{rosin2022time} introduce also another distance method, time-diff, tailored to TempoBERT and we also report its results for comparison.

\item \textbf{Sentence time prediction:} 
 Unlike document dating or event occurrence time prediction which use either long articles or event descriptions as input, sentence time prediction task assumes predicting the writing time of short sentences.
Same as TempoBERT, we utilize here the NYT-years dataset with 40 classes corresponding to 40 years from 1981 to 2020 with 10k sentences sampled per each year.  
Accuracy is used as a metric. Since 20 years are overlapping with the corpus that we used for pre-training of BiTimeBERT, we additionally report the accuracy scores within and outside the overlapped time period (i.e., 1987-2007, 1981-1986 \& 2008-2020). 
All models are fine-tuned for 10 epochs, with a learning rate 2e-05.
\end{enumerate}



Table~\ref{temp_semantic_change} presents the results of semantic change detection. When considering both datasets using cos\_dist, BiTimeBERT achieves the best results with significant correlations ($p < 0.005$), especially on SemEval-Eng which is a long-term corpora. In addition, we can see that TempoBERT using its tailored time-diff method outperforms BiTimeBERT and obtains the best performance on LiverpoolFC of short time spans. However, compared with BiTimeBERT which achieves relatively good results on different types of corpora using the same cos\_dist method, there is a large performance degradation on SemEval-Eng when using time-diff. 
Thus, one needs to be careful when using TempoBERT on semantic change detection, as the corpora type (long-term or short-term) should be known in advance to use an appropriate measuring method (cos\_dist or time-diff).

Table~\ref{tab_SentTimePred} presents the results of sentence time prediction. When considering NYT-years of entire 1987-2020, we can see that TempoBERT is surpassed by BERT, while BiTimeBERT outperforms all other models by a large margin. 
Moreover, for the time period outside the one of BiTimeBERT's pre-training corpus, i.e., 1981-1986 \& 2008-2020, BiTimeBERT outperforms BERT and BERT-NYT with 10.89\%, 19.40\% improvement, respectively. Therefore, despite the fact that TempoBERT has been specifically designed for semantic change detection, our proposed BiTimeBERT can also obtain good results on sentence time prediction.


\begin{table}
\begin{center}
\footnotesize
\centering \caption{Sentence time prediction results.}
\vspace{-1.8em}
\label{tab_SentTimePred}
\renewcommand{\arraystretch}{1.0}
\setlength{\tabcolsep}{0.6em}
\centering
\begin{tabular}{|p{1.7cm}|c|c|c|}
\hline
    \multirow{3}{3cm}{\textbf{Model}} & \multicolumn{3}{c|}{\textbf{NYT-years}} \\
    \cline{2-4}
     & \multicolumn{1}{c|}{\textbf{1981-2020}} & \multicolumn{1}{c|}{\textbf{1987-2007}} & \multicolumn{1}{c|}{\textbf{1981-1986 \& 2008-2020}} \\
    \cline{2-4}
     & \textbf{ACC} & \textbf{ACC}  & \textbf{ACC}   \\
    \hline
    \makecell[l]{BERT} & 10.02  & 9.7 & 10.38 \\
    \hline
    \makecell[l]{BERT-NYT} & 10.23 & 10.75 & 9.64\\
    \hline
    TempoBERT \cite{rosin2022time} & 9.24 & - & - \\
    \hline
    \makecell[l]{BiTimeBERT} & \textbf{12.52}  & \textbf{13.44} & \textbf{11.51} \\
    \hline
  \end{tabular}
\end{center}
\vspace{-2.0em} 
\end{table}

\vspace{-0.7em}
\section{Conclusions} 
In this paper, we have presented a novel and effective language representation model called BiTimeBERT designed specifically for time-related tasks. BiTimeBERT is trained over a temporal news collection through two new pre-training tasks that involve two kinds of temporal aspects (timestamp and content time). We next conducted extensive experiments to investigate the effectiveness of the proposed pre-training tasks. The results reveal that BiTimeBERT can offer effective time-aware representations and could help achieve improved performance on various time-related downstream tasks. 
In the future, 
we will investigate how to incorporate TAMLM with TIR, as both these objectives utilize the same temporal information extracted from content.


\vspace{-0.5em}
\section{Acknowledgments}

This work was supported by the National Natural Science Foundation of China (62076100), Fundamental Research Funds for the Central Universities, SCUT (x2rjD2220050), the Science and Technology Planning Project of Guangdong Province (2020B0101100002).

\FloatBarrier
\bibliographystyle{ACM-Reference-Format}
\bibliography{BiTimeBERT}
\FloatBarrier

\appendix

\end{document}